# DuEDL: Dual-Branch Evidential Deep Learning for Scribble-Supervised Medical Image Segmentation

Yitong Yang[1], Xinli Xu[1,(✉)], Haigen Hu[1], Haixia Long[1], Qianwei Zhou[1] and Qiu Guan[1]

[1]College of Computer Science and Technology,
Zhejiang University of Technology, Hangzhou, China
(✉)`xxl@zjut.edu.cn`

**Abstract.** Despite the recent progress in medical image segmentation with scribble-based annotations, the segmentation results of most models are still not robust and generalizable enough in open environments. Evidential deep learning (EDL) has recently been proposed as a promising solution to model predictive uncertainty and improve the reliability of medical image segmentation. However directly applying EDL to scribble-supervised medical image segmentation faces a tradeoff between accuracy and reliability. To address the challenge, we propose a novel framework called Dual-Branch Evidential Deep Learning (DuEDL). Firstly, the decoder of the segmentation network is changed to two different branches, and the evidence of the two branches is fused to generate high-quality pseudo-labels. Then the framework applies partial evidence loss and two-branch consistent loss for joint training of the model to adapt to the scribble supervision learning. The proposed method was tested on two cardiac datasets: ACDC and MSCMRseg. The results show that our method significantly enhances the reliability and generalization ability of the model without sacrificing accuracy, outperforming state-of-the-art baselines. The code is available at https://github.com/Gardnery/DuEDL.

## 1 Introduction

The goal of medical image segmentation is to clearly capture the anatomical or pathological structural changes in the image, so as to improve the accuracy and efficiency of segmentation, and more effectively assist doctors in clinical diagnosis and analysis. In recent years, convolutional neural networks (CNNs) and Transformers have achieved remarkable results in medical image segmentation. However, most of these methods require precisely labeled datasets at the pixel level for training. Although accurate mask labels are crucial for the generalization ability of the model, their annotation requires a lot of time costs from professional medical experts.

In order to alleviate the time cost pressure caused by accurate annotation, two strategies are proposed. One is to use semi-supervised learning (SSL) to train the model by jointly using a large number of unlabeled data and a moderate number of accurately labeled data. The other is to use weakly supervised learning (WSL) to use sparsely labeled data sets, such as points, boxes, and scribbles, instead of accurate labels. Scribble-supervised are highly prized for their user-friendly interaction [1].

Although Scribble-Supervised methods have made significant progress in improving model performance, current research mainly focuses on optimizing model performance. For example, recent works [2][3][4][5]pay more attention to improving the Dice scores of models. Very little work has been devoted specifically to investigating the robustness and generalization ability of models.

Uncertainty estimation is considered as the key to reflect the reliability of the model, which can be used to quantify the probability of the model being wrong in the prediction. One of the classical methods for uncertainty estimation in Bayesian neural networks is Monte Carlo Dropout (MC Dropout) [6]. However, this method will bring expensive computational cost, and the network will confuse arbitrary uncertainty and epistemic uncertainty for prediction [6][7]. Sensoy et al. proposed a new uncertainty assessment method, i.e., evidential deep learning (EDL), in which the output of the network is regarded as the evidence parameterized by Dirichlet distribution according to Dempster-Shafer evidence theory [8][9] and subjective logic [10]. This method outputs for each object the quality of beliefs for different classes as well as an overall uncertainty value [11], showing impressive performance in terms of model judgment and uncertainty quantification.

Existing EDL-based uncertainty quantification methods are usually only used in image segmentation tasks with accurately labeled datasets. Inspired by literature [4], we propose a scribble-supervised medical image segmentation method based on Dual-Branch Evidential Deep Learning (DuEDL). To the best of our knowledge, we are the first to utilize EDL in scribble-supervised medical image segmentation. The main contributions can be summarized as follows:

- We propose a framework named Dual-Branch Evidential Deep Learning (DuEDL) for scribble-supervised medical image segmentation. We extend the decoder of the segmentation network into two separate branches, and then fuse the evidence obtained from the two branches to generate high-quality pseudo-labels for joint training.
- We improve the joint loss for scribble-supervision in medical image segmentation. We propose a parial evidence loss to adjust the model's evidence learning from the scribble annotations. We utilize the hard pseudo labels from the fused prediction probability of Dirichlet distribution to calculate the consistency loss for the dual-branch evidence.
- Extensive experiments on the two publicly available cardiac MRI segmentation dataset (ACDC and MSCMRseg), demonstrate the better robustness and generalization abilities of the proposed method than those of the existing methods in out-of-distribution (OOD) segmentation.

## 2 Related Work

### 2.1 Scribble-Supervised Image Segmentation

Scribble-supervised image segmentation leverages scribble-based annotations to train models, balancing cost-effectiveness with a pursuit of precision. Researchers like Lee et al. (2020) improve the segmentation performance of models by using EMA to filter the saved segmentation results during the iterative process to obtain reliable pseudo-labels[12]. Ji et al. (2019) employ multiple networks to broaden label datasets. However, these strategies potentially introduce noisy data [13].To improve segmentation, some scholars apply data augmentation and regularization techniques; for example, Zhang et al. (2022) enhance data robustness through mixed augmentation, applying cycle consistency for training model regularization[3]. Liu et al. (2022) use an uncertainty-aware average teacher framework to guide the student model [2]. In recent advancements, Li et al. (2023) fuse CNNs with Transformers to extract refined visual features. Their proposed method for scribble-supervised medical image segmentation capitalizes on multi-modal information enhancement and class embedding to further aid supervision [5].

However, these works lack the analysis of the robustness and generalization ability of the model, while DuEDL aims to enhance the robustness and generalization ability of the model while improving the segmentation accuracy.

### 2.2 Evidential Deep Learning

Evidential deep learning (EDL) is first proposed by Sensoy (2018) et al. [11], which parameterizes the Dirichlet concentration distribution according to Dempster-Shafer theory of evidence (DST) [8][9] and subjective logic [10], replacing the method of using Softmax function to map the output layer to a probability distribution.

Some researchers directly use EDL to improve the performance of models, and researchers like Hemmer et al. (2020) annotate unlabeled data with uncertainty evidence obtained by the Dirichlet distribution [14]. Zou et al. (2022) build upon this groundwork by assigning *belief mass* and *uncertainty mass* to each voxel, thus enhancing the reliability of segmentation outcomes [15]. In order to enhance EDL capability in quantifying uncertainty, Park et al. (2023) redefined the methodology for generating evidence and developed specialized evidence strategies to optimize performance in model detection[16]. Li et al. (2023) designed an evidence loss function to minimize the prediction error of regions on the basis of [15]. It further improves the segmentation performance of the model on brain tumors

Inspired by these approaches, we propose the new framework by combining scribble-supervised medical image segmentation with evidential deep learning

### 2.3 Evidence Fusion

The Dempster-Shafer theory [9] offers a mathematical framework for decision-making and inference under uncertainty by combining evidence from various sources.

Some previous work explored evidence fusion methods. One approach is the evidence integration of neurons. Tong et al. (2021) propose a hybrid architecture E-FCN [18] consisting of a fully convolutional network and evidential neural network based on the method of [19]. Another method is to fuse the evidence obtained from different modalities by evidence fusion rules [20]. For example, Xu et al. (2022) use prior knowledge to assist multi-modal evidence fusion [21]. Shao et al. (2024) adopt two BBA methods to eliminate the bias caused by a single BBA method and propose a multi-modal fusion framework [22].

Unlike the above works, we use the dual-branch framework to fuse the evidence of the single modality to obtain more reliable evidence to improve the segmentation accuracy of the model.

## 3 Method

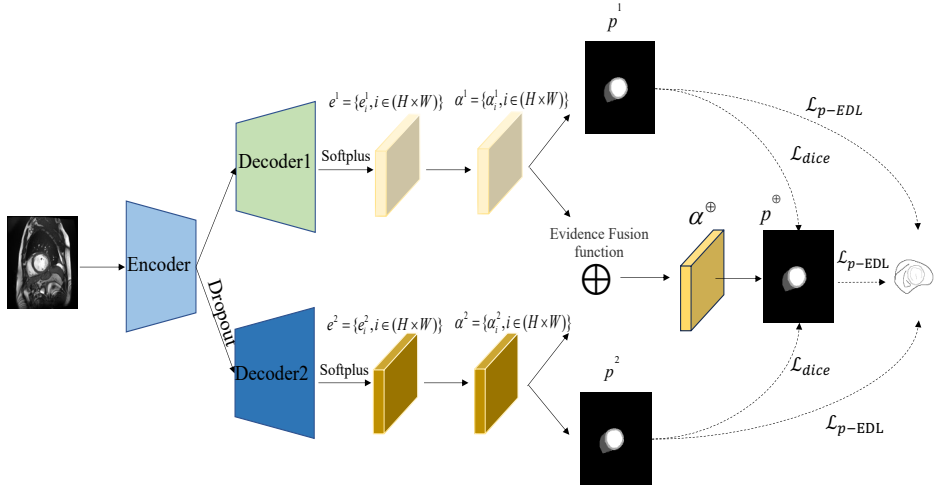

**Fig. 1.** Overview of Dual-Branch Evidential Deep Learning (DuEDL). The class-wise evidence $e^1$ and $e^2$ obtained from the Decoder1 and Decoder2, respectively. A partial evidence loss $L_{p\text{-}EDL}$ is proposed to guide the learning of evidence under the scribble annotations. At the same time, the Dirichlet distribution parameters ($\alpha^1$ and $\alpha^2$) of the first and second branches are also fused to get the parameter $\alpha^\oplus$, and then the hard pseudo-label is generated by the fused Dirichlet prediction probability $p^\oplus$ to guide the consistent evidence of the upper and lower branches.

The proposed framework called DuEDL is illustrated in Fig. 1, which consists of three components: a shared Encoder, the main Decoder1 and an auxiliary Decoder2. For example, given a cardiac MRI image as input, we replace the Softmax function of the decoder with an activation function layer (e.g., Softplus) to ensure non-negative outputs that represent class-wise evidence. The evidence can formulate a Dirichlet distribution so that the multi-class probabilities and predictive uncertainty of each pixel in the input image can be determined. We utilize the fused Dirichlet prediction probabilities from the evidence of two branches to generate hard pseudo labels that assist the network training. Inference of the main Decoder1 in the open medical environments with low quality or out-of-distribution (OOD) can output the segmentation results and uncertainty results to aid judgment.

### 3.1 Pixel Evidence collection

Based on Dempster-Shafer evidence theory (DST) [8][9] and Subjective logic (SL) [10], evidential deep learning (EDL) can quantify the uncertainty of an image segmentation network (including an encoder and a decoder). EDL uses the Dirichlet distribution to model the predicted category distribution for each pixel in the medical image. In EDL, each pixel is assigned *belief mass* and *uncertainty mass*. The belief mass $b$ represents the probability of evidence assigned to each category, while the uncertainty mass $u$ provides uncertainty estimation indicating a lack of evidence [23].

In the $K$-classified medical image segmentation, for the $i$-th pixel, the belief mass $b_{i,j}$ and uncertainty $u_i$ are computed as:

$$b_{i,j} = \frac{e_{i,j}}{S_i} = \frac{\alpha_{i,j} - 1}{S_i} \quad \text{and} \quad u_i = \frac{K}{S_i}, \tag{1}$$

with the restrictions that $\sum_{j=1}^{K} b_{i,j} + u_i = 1$, where $u_i \geq 0, b_{i,j} \geq 0$, $i \in [1, H \times W]$, $j \in [1, K]$, $H$ and $W$ represent the height and width of the input image $x$, respectively. The evidence is $e = \{e_{i,j} \mid i = 1, 2, ..., H \times W, j = 1, 2, ..., K\}$, while $e_{i,j}$ represents the quantity of supporting observations collected from data in favor of the $i$-th pixel be classified into the $j$-th class. Assuming a non-informative prior, the Dirichlet distribution follows a uniform distribution and its concentration parameter $\alpha_{i,j} = e_{i,j} + 1$. The total strength of the Dirichlet distribution is denoted as $S_i = \sum_{j=1}^{K} \alpha_{i,j}$, with higher $S_i$ indicating greater confidence in the distribution [23].

The predicted category probability distribution of the $i$-th pixel is given by:

$$p_{i,j} = \frac{\alpha_{i,j}}{S_i}. \tag{2}$$

The proposed framework utilizes the Softplus function to transform the output of the segmentation network into non-negativity evidence for each pixel. Specifically, the evidence of the first and second branches can be represented as $e^n = \{e_{i,j}^n \mid i = 1, 2, ..., H \times W, j = 1, 2, ..., K, n = 1, 2\}$, where $n=1$ and $n=2$ represent the evidence from the upper and lower branch, respectively. Here, the decoder parameters of the second branch are the same as those of the first branch's decoder, and the encoder result after dropout is used as the input to the decoder of the second branch, which can reduce the model's parameter training load. The first and second branches output different evidence. The Dirichlet concentration parameters $\alpha^n = \{\alpha_{i,j}^n \mid i=1,2,...,H \times W, j=1,2,...,K, n=1,2\}$ can be further computed as $\alpha^n = e^n + 1$.

## 3.2 Partial Evidence Learning

Sensoy et al. found that using a combination of Expected Mean Square Error (MSE) and Kullback-Leibler (KL) divergence is more stable for training evidence neural networks than cross-entropy loss and negative log-likelihood, resulting in better performance [11]. However, when dealing with scribble-annotation, where only a few pixels are labeled and most pixels are unlabeled, directly using EDL loss can lead to learning incorrect information and increased uncertainty in predictions for unlabeled pixel areas.

We propose the partial evidence learning loss $L_{p-EDL}$ combined with the partial Mean Square Error loss $L_{pMSE}$ and the partial Kullback-Leibler loss $L_{pKL}$ only for annotated pixels:

$$L_{p-EDL}(p, y) = L_{pMSE}(p, y) + \lambda_t L_{pKL}(Dir(p \mid \hat{\alpha}) \parallel Dir(p \mid 1)) \quad (3)$$

Where $\lambda_t = \min(1, \frac{t}{\beta})$ represents annealing coefficient, $t$ denotes the current iteration step, and $\beta$ denotes a hyperparameter.

The partial MSE loss $L_{pMSE}$ is computed as follows:

$$L_{pMSE}(p, y) = \sum_{i=1}^{|\omega_y|} \sum_{j=1}^{K} (y_{i,j} - p_{i,j})^2 + \frac{p_{i,j}(1 - p_{i,j})}{S_i + 1} \quad . \quad (4)$$

Where $y$ denotes the ground truth, $|\omega_y|$ represents the number of labeled pixels in the input mage. $y_{i,j}$ and $p_{i,j}$ are the label and predicted probability of the $i$-th pixel for class $j$. $S_i$ denotes the strength of the Dirichlet distribution for the $i$-th pixel.

The partial Kullback-Leibler function $L_{pKL}$ can be further calculated as follows:

$$L_{pKL}(Dir(p \parallel \hat{\alpha}) \parallel Dir(p \mid 1)) = \sum_{i=1}^{|\omega_y|} \log \frac{\Gamma\left(\sum_{j=1}^{K} \hat{\alpha}_{i,j}\right)}{\Gamma(K) \prod_{j=1}^{K} \Gamma(\hat{\alpha}_{i,j})} + \sum_{j=1}^{K} (\hat{\alpha}_{i,j} - 1)\left[\psi(\hat{\alpha}_{i,j}) - \psi\left(\sum_{j=1}^{K} \hat{\alpha}_{i,j}\right)\right] \quad (5)$$

Where $Dir(p \mid \hat{\alpha})$ denotes the Dirichlet distribution of the predicted probability $p$ after removing non-informative Dirichlet distribution parameters, where $\hat{\alpha} = \alpha \odot (1 - y) + y$, while $\alpha = e + 1$ represents the Dirichlet distribution parameter for the $i$-th pixel, and $y_i$ is the label in one-hot encoding for the $i$-th pixel. $Dir(p \mid 1)$ denotes the uniform Dirichlet distribution with all parameters set to 1. $\Gamma()$ and $\psi()$ represents the gamma function and the digamma function, respectively.

The partial evidence loss reduces the possibility of the model learning incorrect information and can utilize limited supervision effectively. Therefore, it can improve the prediction accuracy and uncertainty estimation of the segmentation model for unlabeled pixel regions under scribble-supervision.

### 3.3 Dual-Branch Evidence Fusion

Most evidence fusion strategies are based on multimodal fusion or neural network ensembles to obtain reliable evidence. The evidence of the first and second branches is denoted as $e^1$ and $e^2$, respectively. The aiming of evidence fusion is to eliminate potential allocation biases caused by a single branch. According to Eq.1 in section 3.1, we can obtain the belief mass of dual branches $b^n = \{b_{i,j}^n | i=1,2,...,H \times W, j=1,2,...,K, n=1,2\}$ and the uncertainty of dual branches $u^n = \{u_i^n | i=1,2,...,H \times W, n=1,2\}$. Dempster-Shafer evidence theory (DST) provides the fusion of the belief mass and uncertainty as follow:

$$b_{i,j}^{\oplus} = \frac{1}{1-C}(b_{i,j}^1 b_{i,j}^2 + b_{i,j}^1 u_i^2 + b_{i,j}^2 u_i^1), u_i^{\oplus} = \frac{1}{1-C} u_i^1 u_i^2 \quad \text{and} \quad C = \sum_{n \neq m} b_{i,n}^1 b_{i,m}^2. \tag{6}$$

The fused belief mass $b^{\oplus} = \{b_{i,j}^{\oplus} | i=1,2,...,H \times W, j=1,2,...,K\}$ and uncertainty $u^{\oplus} = \{u_i^{\oplus} | i=1,2,...,H \times W\}$ are utilized to calculate the fused evidence, strength and parameter of Dirichlet distribution:

$$e_{i,j}^{\oplus} = K \frac{b_{i,j}^{\oplus}}{u_i^{\oplus}} \langle S_i^{\oplus} = \frac{K}{u_i^{\oplus}} \quad \text{and} \quad \alpha_{i,j}^{\oplus} = e_{i,j}^{\oplus} + 1. \tag{7}$$

The above fusion strategy allows the proposed model to better utilize evidence from different sources. The fused strength and parameter of Dirichlet distribution can be used to calculate the fused prediction probability $p^{\oplus} = \{p_{i,j}^{\oplus} | i=1,2,...,H \times W, j=1,2,...,K\}$ according to Eq.2.

We apply $p^{\oplus}$ to generate hard pseudo labels $\bar{p}^{\oplus} = \arg\max(p^{\oplus})$ for unlabeled pixels, inspired by dynamically mixed pseudo labels supervision [4]. The advantage of hard pseudo labels is to cut off the gradient between the upper and lower branches to maintain their independence [4], we use hard pseudo labels to calculate the consistency loss $L_{ECL}$ of the upper and lower evidence:

$$L_{ECL} = 0.5(L_{dice}(\bar{p}^{\oplus}, p^1) + L_{dice}(\bar{p}^{\oplus}, p^2)). \tag{8}$$

Where $L_{dice}$ is the Dice loss. $p^1 = \{p_{i,j}^1 | i=1,2,...,H \times W, j=1,2,...,K\}$ and $p^2 = \{p_{i,j}^2 | i=1,2,...,H \times W, j=1,2,...,K\}$ are the prediction probabilities of Dirichlet distributions for the first and second branches according to Eq.2, respectively.

The consistency loss is applied to both the upper and lower branches, so that better pseudo-supervision information can be propagated to the unlabeled pixels, thereby improving the generalization ability and robustness of the segmentation model.

### 3.4 Jointly Optimize Loss

The final jointly loss $L_{total}$ is divided into two parts: the supervised loss $L_S$, and the unsupervised evidence consistency loss $L_{ECL}$:

$$L_{total} = L_s + \lambda_u L_{ECL}. \tag{9}$$

Where $\lambda_u$ is the weight for the unsupervised loss $L_{ECL}$. The supervised loss $L_s$ includes three items:

$$L_s = \frac{1}{3}(\sum_{n=1}^{2} L_{p-EDL}(\boldsymbol{p}^n, \boldsymbol{y}) + L_{p-EDL}(\boldsymbol{p}^{\oplus}, \boldsymbol{y})). \tag{10}$$

Where $L_{p\text{-}EDL}$ is the partial evidence loss function. $n=1$ and n=2 represents the upper and lower branch, respectively.

Finally, the proposed network can be trained with scribble annotations by minimizing the above joint object function $L_{total}$.

## 4 Experiments

In this section, we evaluate DuEDL on the two different cardiac datasets, including the segmentation performance, robustness, and generalization ability. Meanwhile, we evaluate the effectiveness of different components of DuEDL.

### 4.1 Datasets

We select two cardiac MRI image segmentation datasets: ACDC and MSCMRseg, where MSCMRseg as the out-of-distribution dataset. Figure 2 illustrates an example of each dataset.

**ACDC:** The ACDC cardiac dataset [24] consists of cine-MRI images of 100 patients. For each patient's cine-MRI images, the dataset provides precise and shorthand annotations of the right ventricle (RV), left ventricle (LV), and myocardium (MYO) for the diastolic (ED) and systolic (ES) phases of the heart, respectively. In the experiment, 70 patient data are randomly divided as the training set, 15 patient data as the validation set, and 15 patient data as the test set.

**MSCMRseg:** The MSCMRseg dataset [25][26] includes late gadolinium-enhanced (LGE) MRI images of 45 patients with cardiomyopathy. Each patient's data has both precise and simplified annotations of LV, MYO, and RV.

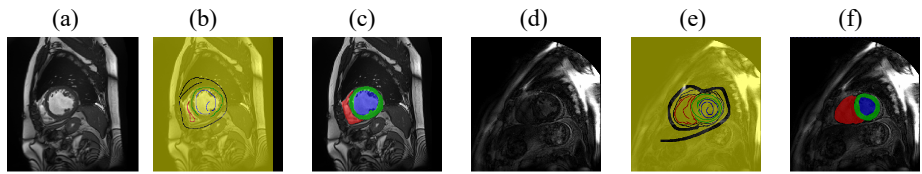

**Fig. 2**. Examples of datasets ACDC and MSCMRseg. (a) image, (b) scirble annotation and (c) ground truth are from ACDC. (d) image, (e) scirble annotation and (f) ground truth are from MSCMRseg.

### 4.2 Evaluation Metrics

We employ four widely-used metrics, such as Dice, Average Symmetric Surface Distance (ASSD) [27], expected calibration error (ECE) [28], and uncertain-error overlap (UEO) [28]. Dice and ASSD are commonly used to evaluate the segmentation performance of the model. ECE and UEO are used to evaluate the performance of the model in uncertainty estimation.

### 4.3 Implementation Details

The proposed framework uses UNet[29] with encoder-decoder structure as the backbone, and expands the decoder of UNet into two slightly different branches. The second branch decoder shares parameters with the first branch decoder except that the second branch decoder adds a dropout layer before the input (where drop rate ratio=0.5). The network model is implemented using Pytorch and trained on an NVIDIA GeForce RTX 3090 Ti.

We refer to the preprocessing method and the parameters setting in [3]: each slice of the dataset is intensity-scaled to the range of 0 to 1. We use random rotation, and random flip to generate the training set, and resize the augmented picture to 256×256 as the input of the network. Stochastic Gradient Descent (SGD) is employed to minimize the joint loss function, with a weight decay rate set to $10^{-4}$ and momentum set to 0.9. We set the number of epochs for training to 300. Other parameter settings include setting in Eq.3 to 10 and setting in Eq.10 to 0.3.

### 4.4 Results

We evaluate the segmentation performance of our proposed DuEDL model on the ACDC dataset, comparing it with other different methods: the scribble-supervised medical image segmentation methods (WSL) and the fully supervised methods (FSL). The main WSL methods include : 1) DMPLS [4], a dual-branch method using dynamically mixed pseudo labels supervision, 2) USTM [2], a uncertaintyaware self-ensembling and transformation-consistent model, 3) CycleMix [3], a segmentation model based on hybrid enhancement and cycle consistency , and 4) ScribbleVC [5], a segmentation model based on visual class embedding. In addition, we incorporate the evidence loss function from TBraTS [17] and R-BEDL [18] into the dual-branch scribble-supervised segmentation framework [3], resulting in $DMPLS_T$ and $DMPLS_R$ TBraTS [15] is a confidence brain tumor segmentation model, while R-BEDL [17] is a brain tumor segmentation model based on region-based evidential deep learning.

Table 1. The proposed method is compared with other advanced methods on the ACDC dataset. Bold indicates the best performance among WSL.

| Type | Method | Dice | ASSD | ECE | UEO |
|---|---|---|---|---|---|
| WSL | USTM | 0.745 | 13.32 | 0.061 | 0.611 |
| | CycleMix | 0.848 | 6.02 | 0.052 | 0.752 |
| | ScribbleVC | **0.881** | 2.71 | **0.003** | 0.752 |
| | DMPLS | 0.846 | 5.76 | 0.119 | 0.788 |
| | $DMPLS_T$ | 0.849 | 5.59 | 0.101 | 0.768 |
| | $DMPLS_R$ | 0.847 | 4.19 | 0.129 | 0.798 |
| | DuEDL | 0.875 | **2.34** | 0.014 | **0.802** |
| FLS | $DMPLS_F$ | 0.912 | 1.84 | 0.069 | 0.852 |

The segmentation performance comparison between our model and the mentioned methods on the ACDC dataset is shown in Table 1. $DMPLS_F$ refers to the DMPLS model [4] trained with the fully supervised labels (i.e., Ground Truth). In terms of segmentation performance, DuEDL is comparable to ScribbleVC, ranking just below

DMPLS$_F$. It achieves the best performance among the medical image segmentation methods based on evidential deep learning, obtaining the highest Dice value and the smallest ASSD value. Compared to the latest scribble-supervised segmentation methods, DuEDL has a second-highest Dice value after ScribbleVC, while its ASSD value surpasses ScribbleVC, ranking first. This indicates that DuEDL effectively maintains segmentation performance in scribble-supervised learning. In terms of uncertainty evaluation, DuEDL ranks second in ECE value among all methods, indicating good error calibration in the proposed method. Apart from the fully supervised method, DuEDL has the highest UEO value among all scribble-supervised methods, indicating its ability to more accurately localize errors and estimate uncertainty. The comparison results in Table 1 effectively validate the effectiveness and superiority of DuEDL in the scribble-supervised cardiac segmentation.

To validate the robustness of DuEDL, we use Gaussian filters to blur the ACDC dataset following the experimental setup in reference [17], which simulates the low-quality data acquisition scenarios. We use 3 different Gaussian filters with standard deviations $\sigma$ of 0.05, 0.1 and 0.15, respectively.

As shown as Table 2, when $\sigma$ =0.05, DuEDL achieves the lowest ASSD value among all methods, while its Dice value (0.862) and UEO value (0.785) are also superior to other scribble-supervised segmentation methods except for the fully supervised DMPLS$_F$ (Dice=0.875, UEO=0.821). With increasing $\sigma$, all methods experience a decrease in segmentation performance; however, our method still outperforms others in terms of Dice, ASSD, and UEO values and even surpasses the fully supervised DMPLS$_F$. Compared to DMPLS$_F$, when $\sigma$ =0.1 and $\sigma$ =0.15, DuEDL shows an improvement of 0.171 and 0.439 in Dice value, 7.16 and 19.32 in ASSD value, as well as 0.237 and 0.406 in UEO value, respectively. Furthermore, DuEDL maintains a stable ECE value as $\sigma$ increases, which consistently remains smaller than that of the fully supervised method, and consistently, ranking first among all methods. Overall, these results indicate that DuEDL exhibits strong robustness and accurate segmentation in scribble annotations even under deteriorated quality of cardiac MRI images.

Table 2. Quantitative comparison of different methods on the ACDC test set. Bold indicates the best performance among WSL.

| Type | Method | $\sigma = 0.05$ | | | | $\sigma = 0.1$ | | | | $\sigma = 0.15$ | | | |
|---|---|---|---|---|---|---|---|---|---|---|---|---|---|
| | | Dice | ASSD | ECE | UEO | Dice | ASSD | ECE | UEO | Dice | ASSD | ECE | UEO |
| WLS | USTM | 0.695 | 16.11 | 0.055 | 0.554 | 0.579 | 19.71 | 0.061 | 0.401 | 0.428 | 21.86 | 0.066 | 0.267 |
| | CycleMix | 0.804 | 12.77 | 0.052 | 0.749 | 0.591 | 33.64 | 0.047 | 0.625 | 0.351 | 44.72 | 0.040 | 0.451 |
| | ScribbleVC | 0.641 | 14.28 | **0.011** | 0.581 | 0.055 | 42.27 | 0.032 | 0.209 | 0.001 | 48.92 | 0.035 | 0.148 |
| | DMPLS | 0.757 | 7.15 | 0.133 | 0.715 | 0.472 | 19.77 | 0.123 | 0.434 | 0.096 | 45.34 | 0.094 | 0.159 |
| | DMPLS$_T$ | 0.773 | 6.60 | 0.103 | 0.721 | 0.511 | 15.30 | 0.118 | 0.419 | 0.183 | 32.47 | 0.129 | 0.190 |
| | DMPLS$_R$ | 0.716 | 9.92 | 0.102 | 0.661 | 0.356 | 20.99 | 0.131 | 0.369 | 0.086 | 36.92 | 0.134 | 0.161 |
| | DuEDL | **0.862** | **2.68** | 0.014 | **0.785** | **0.797** | **4.03** | **0.014** | **0.704** | **0.649** | **7.93** | **0.016** | **0.567** |
| FSL | DMPLS$_F$ | 0.875 | 2.86 | 0.091 | 0.821 | 0.626 | 11.19 | 0.091 | 0.467 | 0.210 | 27.25 | 0.103 | 0.161 |

Additionally, we evaluate the generalization ability of DuEDL on out-of-distribution datasets, using ACDC as the training dataset and MSCMRseg as the test set. As shown in Table 3, DuEDL achieves the lowest ASSD value (ASSD=5.76), surpassing the fully supervised $DMPLS_F$ (ASSD=5.83). Although its Dice value (Dice=0.582) ranks slightly lower than $DMPLS_F$ (Dice=0.611), compared to other scribble-supervised segmentation methods, DuEDL demonstrates significantly superior segmentation performance. Furthermore, the uncertainty assessment metrics of the proposed method indicate improved generalization ability to out-of-distribution datasets, with the smallest ECE value and the largest UEO value.

**Table 3.** Comparison of the proposed method with other state-of-the-art methods on the out-of-distribution dataset. Bold indicates the best performance among WSL.

| Type | Method | ACDC → MSCMRseg | | | |
|---|---|---|---|---|---|
| | | Dice | ASSD | ECE | UEO |
| WSL | USTM | 0.297 | 23.44 | 0.081 | 0.195 |
| | CycleMix | 0.319 | 14.24 | 0.015 | 0.210 |
| | ScribbleVC | 0.502 | 9.02 | 0.015 | 0.422 |
| | DMPLS | 0.310 | 17.54 | 0.255 | 0.327 |
| | $DMPLS_T$ | 0.315 | 16.98 | 0.156 | 0.342 |
| | $DMPLS_R$ | 0.214 | 27.01 | 0.102 | 0.267 |
| | DuEDL | **0.582** | **5.76** | **0.013** | **0.530** |
| FLS | $DMPLS_F$ | 0.611 | 5.83 | 0.135 | 0.579 |

### 4.5 Ablation Study

To further understand the effectiveness of different strategies in our proposed method, we conduct ablation experiments on strategy 1 (partial evidence optimization loss) and strategy 2 (dual-branch evidence dynamic fusion). Tables 4 and 5 respectively demonstrate the impact of these strategies on the robustness and generalization performance. $Model_1$ refers to the baseline model with a dual-branch architecture (similar to the architecture [3]). $Model_2$ represents the model trained on the baseline using strategy 1 and the dynamic fusion method of the baseline. $Model_3$ is our DuEDL method.

As shown in Table 4, the use of partial evidence optimization loss in Model2 does not improve the model segmentation performance compared to Model1. With increasing noise ($\sigma$), the Dice value decreases for Model2 compared to the baseline Model1. Comparing $Model_2$ and $Model_3$, $Model_3$ shows an improvement in Dice value and a decrease in ASSD value. This indicates that using dual-branch evidence dynamic fusion provides more reliable pixel classification evidence than a single branch, resulting in better-quality pseudo-labels and improved supervision of unlabeled pixels. Furthermore, even with the low-quality data ($\sigma$=0.1), $Model_3$ performs better with an improved Dice value of 0.797 and ASSD value of 0.403.

Table 5 shows that the use of partial evidence fusion in Model2 improves segmentation performance on out-of-distribution data (Dice value increases by 0.162 compared to Model1). However, there is no improvement in ASSD; in fact, ASSD increases. Nevertheless, it enhances robustness (ECE value decreases by 0.196, UEO value increases by 0.072) compared to Model$_1$. Additionally, incorporating dual-branch evidence dynamic fusion further enhances uncertainty evaluation for out-of-distribution segmentation in Model$_3$. Model$_3$ exhibits improved segmentation performance (Dice value increases by 0.11 compared to Model$_2$, ASSD value decreases by 20.72) and enhanced robustness (ECE value decreases by 0.046, UEO value increases by 0.131) compared to Model$_2$.

Table 4. The impact of different strategies on robustness. Bold indicates the best performance.

| Method | $\sigma = 0$ | | | | $\sigma = 0.05$ | | | | $\sigma = 0.1$ | | | |
|---|---|---|---|---|---|---|---|---|---|---|---|---|
| | Dice | ASSD | ECE | UEO | Dice | ASSD | ECE | UEO | Dice | ASSD | ECE | UEO |
| Model$_1$ | 0.846 | 5.76 | 0.119 | 0.788 | 0.757 | 7.15 | 0.133 | 0.715 | 0.472 | 19.77 | 0.123 | 0.434 |
| Model$_2$ | 0.841 | 16.01 | **0.011** | 0.746 | 0.813 | 16.29 | **0.011** | 0.722 | 0.706 | 18.45 | **0.013** | 0.591 |
| Model$_3$ | **0.875** | **2.34** | 0.014 | **0.802** | **0.862** | **2.68** | 0.014 | **0.785** | **0.797** | **4.03** | 0.014 | **0.704** |

Table 5. The impact of different strategies on generalization ability. Bold indicates the best performance.

| ACDC → MSCMRseg | | | | |
|---|---|---|---|---|
| Method | Dice | ASSD | ECE | UEO |
| Model$_1$ | 0.310 | 17.54 | 0.255 | 0.327 |
| Model$_2$ | 0.472 | 26.48 | 0.059 | 0.399 |
| Model$_3$ | **0.582** | **5.76** | **0.013** | **0.530** |

## 5 Conclusions

In this work, we propose DuEDL to address the issues in scribble-supervised medical image segmentation. We introduces partial evidence loss and dual-branch evidence fusion. Specifically, the partial evidence loss aims to optimize evidence learning for the labeled pixels, while the dual-branch evidence fusion ensures consistent learning for unlabeled pixel evidence. Experimental results on public cardiac datasets ACDC and MSCMRseg demonstrate that our DuEDL method outperforms existing state-of-the-art methods in terms of robustness and generalization ability for scribble-supervised segmentation. Moreover, DuEDL has a better trade-off between robustness and generalization ability and segmentation accuracy than other medical image segmentation methods based on evidential deep learning. Future work will focus on further improving joint optimization strategies to maintain excellent reliability while achieving high-performance segmentation.

**Acknowledgments**. This work was jointly supported by the National Natural Science Foundation of China (Grant no. 62373324, 62176236, 62271448 and U20A20171), the Zhejiang Provincial Natural Science Foundation of China (Grant no. LY23F030008 and LGF22F030016) and the Key Research and Development Program of Zhejiang Province (Grant no. 2022C03113).